\documentclass[journal,twoside,web]{ieeecolor}
\usepackage{generic}
\usepackage{cite}
\usepackage{amsmath,amssymb,amsfonts}
\usepackage{algorithmic}
\usepackage{graphicx}
\usepackage{algorithm,algorithmic}
\usepackage{hyperref} 
\hypersetup{hidelinks=true}
\usepackage{textcomp} 
\usepackage{booktabs}  
\usepackage{multicol}  
\usepackage{multirow}
\usepackage{pifont} 
\usepackage{bbding}
\usepackage{xcolor}
\usepackage[table]{xcolor}

\def\BibTeX{{\rm B\kern-.05em{\sc i\kern-.025em b}\kern-.08em
    T\kern-.1667em\lower.7ex\hbox{E}\kern-.125emX}}
\begin{document}
\title{Prior-guided Hierarchical Instance–pixel Contrastive Learning for Ultrasound Speckle Noise Suppression} 
\author{Zhenyu Bu, Yuanxin Xie, and Guang-Quan Zhou, \IEEEmembership{Senior Member, IEEE}
\thanks{Manuscript submitted Feb, 2026. This work was supported by the National Natural
Science Foundation of China under Grant 62371121. (Corresponding author: Guang-Quan Zhou.)}
\thanks{Zhenyu Bu is with the School of Biological Science and Medical Engineering, Southeast University, Nanjing 211189, China, and with Department of Biomedical Engineering, The Ohio State University, Columbus 43210, USA. (e-mail: zhenyu.bu@osumc.edu). }
\thanks{Yuanxin Xie is with the School of Biological Science and Medical Engineering, Southeast University, Nanjing 211189, China (e-mail: 230258624@seu.edu.cn).}
\thanks{Guang-Quan Zhou is with the School of Biological Science and Medical Engineering, Southeast University, Nanjing 211189, China, and with
the Jiangsu Key Laboratory of Biomaterials and Devices, Southeast University, Nanjing 211189, China, and also with the State Key Laboratory
of Digital Medical Engineering, Southeast University, Nanjing 211189,
China (e-mail: guangquan.zhou@seu.edu.cn).}}

\maketitle

\begin{abstract}
 Ultrasound denoising is essential for mitigating speckle-induced degradations, thereby enhancing image quality and improving diagnostic reliability. Nevertheless, because speckle patterns inherently encode both texture and fine anatomical details, effectively suppressing noise while preserving structural fidelity remains a significant challenge. In this study, we propose a prior-guided hierarchical instance–pixel contrastive learning model for ultrasound denoising, designed to promote noise-invariant and structure-aware feature representations by maximizing the separability between noisy and clean samples at both pixel and instance levels. Specifically, a statistics-guided pixel-level contrastive learning strategy is introduced to enhance distributional discrepancies between noisy and clean pixels, thereby improving local structural consistency. Concurrently, a memory bank is employed to facilitate instance-level contrastive learning in the feature space, encouraging representations that more faithfully approximate the underlying data distribution. Furthermore, a hybrid Transformer–CNN architecture is adopted, coupling a Transformer-based encoder for global context modeling with a CNN-based decoder optimized for fine-grained anatomical structure restoration, thus enabling complementary exploitation of long-range dependencies and local texture details. Extensive evaluations on two publicly available ultrasound datasets demonstrate that the proposed model consistently outperforms existing methods, confirming its effectiveness and superiority.
\end{abstract}


\section{Introduction}
\label{sec:introduction}
\IEEEPARstart{U}{ltrasound} imaging is a widely adopted, cost-effective and non-invasive diagnostic modality that plays a critical role across a broad range of clinical applications, including cardiology, obstetrics, and abdominal assessment. 
Compared to other imaging modalities such as CT, MRI, and PET~\cite{duck2020ultrasound, leighton2007ultrasound, zhang2018liver}, ultrasound offers a unique combination of safety, affordability, and real-time imaging capabilities.
It has evolved beyond basic grayscale imaging driven by increasing clinical demand. Technological advancements in ultrasound, including Doppler Imaging~\cite{ho2006clinician} and Elastography~\cite{gennisson2013ultrasound}, have significantly enhanced its diagnostic potential, allowing for the evaluation of blood flow, tissue elasticity and the early detection of conditions such as liver fibrosis. These advantages have established ultrasound as an indispensable modality in both emergency care and preventive healthcare settings.

Despite its clinical advantages, ultrasound imaging is inherently prone to speckle noise caused by the coherent nature of ultrasound wave reflection~\cite{sudha2009speckle}. This artifact severely degrades image clarity, obscures anatomical boundaries, and complicates accurate clinical interpretation~\cite{xie2022improved}. Beyond visual quality, speckle noise also undermines the reliability of downstream tasks such as classification, segmentation and lesion detection~\cite{chen2021lesion, xia20233d}. These challenges underscore the urgent need for robust and structure-aware denoising techniques to restore diagnostic fidelity and enhance computational performance.

Over the past several years, a variety of techniques have been designed for noise reduction. They can be generally categorized into two groups: non-learning methods and learning-based methods. 
Dabov et al.~\cite{dabov2007image} proposed a Block Matching and 3D Collaborative Filtering (BM3D) image denoising, which leverages the principles of non-local similarity and sparse representation to effectively reduce noise while preserving image details. Michal et al.~\cite{aharon2006k} proposed a K-SVD algorithm. This is an iterative method for sparse representation, aiming at finding a "dictionary" matrix such that the input data can be approximated by a few elements in the dictionary. However, these popular non-learning methods usually introduce excessive smoothing effects when applying on ultrasound datasets, resulting in the missing anatomical details and a certain degree of blur image boundary.

Recently, learning-based methods have emerged as powerful alternatives~\cite{wang2022ffcnet, zhou2023tagnet, wang2024sbcnet}.
These approaches are widely employed for noise reduction due to their robust feature extraction capabilities and adaptability to complex noise environments. Their ability to automatically learn and generalize intricate patterns from data not only compensates for the limitations of traditional techniques but also paves the way for improved image restoration performance and enhanced preservation of fine details. Several CNN-based approaches have been specifically developed to tackle this problem. Denoising Convolutional Neural Network (DnCNN) designed by Zhang et al.~\cite{zhang2017beyond} aims to learn and model noise patterns, which is based on residual learning and the clean image is obtained by subtracting the learned noise from the noisy input image. Chen et al.~\cite{chen2017low} introduced a Residual Encoder-Decoder Convolutional Neural Network (RED-CNN) for low-dose CT denoising. The primary concept involves employing an encoder-decoder architecture to train a model capable of transforming a low-dose CT image into a high-dose CT image. Nevertheless, the performance of RED-CNN may be suboptimal when confronted with complex tasks due to its limited capacity to capture global contextual information. Li et al.~\cite{yancheng2023red} established a residual encoder-decoder network based on multi-attention fusion for ultrasound image denoising. This architecture leverages multiple convolution and deconvolution layers, as well as multi-attention fusion blocks to capture both local and global image features. However, these experiments were conducted using additive Gaussian white noise. In contrast, ultrasound speckle noise typically behaves as multiplicative noise, and its statistical characteristics are more consistent with a Rayleigh distribution. This implies that the experiments cannot accurately capture the true noise distribution inherent in ultrasound images.

Due to their exceptional ability to capture global features and deliver superior performance, transformer models such as ViT, Swintransformer, CSwintransformer~\cite{vaswani2017attention, dosovitskiy2020image, liu2021swin, dong2021cswin} have found extensive use in the realm of computer vision. Numerous models leveraging transformer architectures have been developed to effectively reduce noise. Liu et al.~\cite{liang2021swinir} proposed SwinIR for image super-resolution, denoising and compression artifact removal. It includes shadow feature extraction, deep feature selection and high-quality image reconstruction parts. In the deep feature extraction module, the model employs multiple Residual Swin Transformer Blocks (RSTB), which utilize a shifted window-based multi-head self-attention mechanism to effectively capture both local details and global information in the image. Wang et al.~\cite{wang2022uformer} introduced Uformer, a U-shaped pure transformer network for image restoration. It replaces the MLP with Locally-enhanced Feed-Forward (LeFF) to enhance local information. 
However, these methods are primarily designed for generic image restoration and do not explicitly account for the unique statistical and structural characteristics of ultrasound speckle.
Since speckle patterns inherently encode both texture and fine anatomical details, overlooking this property may lead to suboptimal noise suppression, inadvertently attenuating structural information and potentially influencing clinical decision-making.


To effectively suppress speckle noise while preserving texture and fine anatomical details, we introduce a prior-guided hierarchical instance–pixel contrastive learning framework that encourages noise-invariant and structure-aware representations by enhancing the discriminability between noisy and clean samples at both pixel and instance levels. Specifically, a statistics-guided pixel-level contrastive learning strategy is designed to enforce pixel level structural consistency by explicitly modeling the statistical characteristics of speckle noise, thereby preventing the suppression of fine-grained anatomical details. In parallel, a memory-augmented instance-level contrastive learning module operates in the global feature space to align noisy and clean image global representations, mitigating semantic drift induced by accumulated noise perturbations. Furthermore, we adopt a hybrid Transformer–CNN architecture that leverages the Transformer’s ability to model long-range context and the CNN’s strength in preserving fine-grained anatomical details. Overall, our main contributions are summarized as follows:

\begin{itemize}[]  
    \item To suppress speckle noise while preserving texture and fine anatomical details, we propose a prior-guided hierarchical instance–pixel contrastive learning framework that learns noise-robust and structure-aware representations by contrasting noisy and clean samples at both pixel and instance levels.
    \item We adopt a hybrid Transformer–CNN architecture, where global contextual representations are first modeled by a Transformer encoder and subsequently refined by a CNN decoder to recover fine anatomical structures.
    \item Extensive experiments on two public ultrasound datasets, BUSI and CAMUS, demonstrate that our method consistently outperforms existing approaches across different noise levels.
\end{itemize}


\section{Methodology}
\subsection{Architecture Overview }

\begin{figure*}[!t]
\centerline{\includegraphics[width=\textwidth]{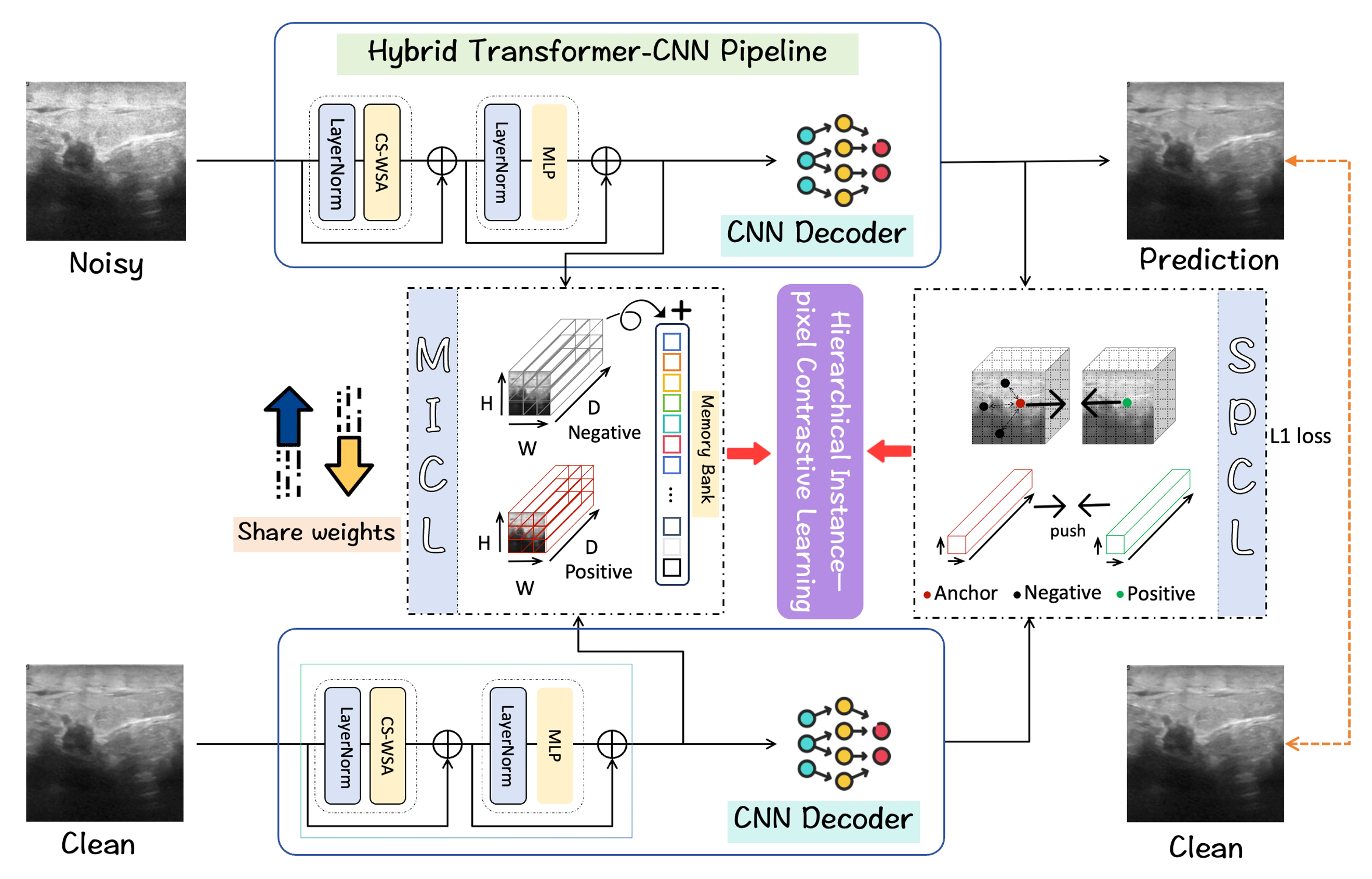}}
\caption{Overall pipeline of the proposed prior-guided hierarchical instance–pixel contrastive learning framework for ultrasound speckle noise suppression. The noisy ultrasound image is encoded by a hybrid Transformer–CNN pipeline, where the Transformer-based encoder captures global contextual features and the CNN-based decoder focuses on fine-grained anatomical structure restoration. Prior-guided hierarchical instance-pixel contrastive learning
module in the middle demonstrates that jointly performing pixel and instance-level contrastive learning to encourage noise-invariant and structure-aware representation.}
\label{framework}
\end{figure*}

The overall architecture is shown in Figure~\ref{framework}. We propose a dual-branch hybrid Transformer-CNN architecture with shared weights, jointly processing both noisy and clean ultrasound images to facilitate representation alignment. Let $I_{\text{noisy}} \in \mathbb{R}^{H \times W}$ denote the noisy input image and $I_{\text{clean}}$ the clean counterpart. We adopt a Gaussian-approximated multiplicative speckle noise model:
\begin{equation}
   I_{\text{noisy}} = I_{\text{clean}} (1 + \varepsilon), \quad \varepsilon \sim \mathcal{N}(0,\sigma^2), 
\end{equation}
which has been widely used as a practical approximation of ultrasound speckle noise.

To unify the representation space for both domains, we define a shared encoder $\Phi_\theta$ that maps inputs from the image manifold $\mathcal{M}$ into a latent token embedding space:
\begin{equation}
\Phi_\theta: \mathcal{M} \rightarrow \mathbb{R}^{N \times d}.
\label{eq:encoder_mapping}
\end{equation}

The input image is partitioned into non-overlapping patches $\{x_i\}_{i=1}^N$ and embedded with positional encodings $p_i$:
\begin{equation}
X_0 = \{E(x_i) + p_i\}_{i=1}^N,
\label{eq:token_input}
\end{equation}
where $E(\cdot)$ denotes patch embedding function.

Inspired by the CSWin Transformer~\cite{dong2021cswin}, these tokens are processed through L CSWin Transformer blocks to improve the model’s efficiency and effectiveness. Each block applies Cross-Shape Window Self-Attention (CS-WSA) and feedforward MLP modules with residual connections:
\begin{equation}
\footnotesize
X_l = X_{l-1} + \text{CS-MSA}(\text{LN}(X_{l-1})) + \text{MLP}(\text{LN}(X_{l-1})), \quad l = 1, \dots, L,
\label{eq:transformer_layer}
\end{equation}
where $\text{LN}(\cdot)$ denotes layer normalization. The encoder output $Z$ is the final token representation:
\begin{equation}
Z = \Phi_\theta(I_{\text{noisy}}) \in \mathbb{R}^{N \times d}.
\label{eq:token_output}
\end{equation}

The decoder $\Psi_\phi$ reshapes and decodes the latent tokens into a full-resolution image:
\begin{equation}
\hat{I} = \Psi_\phi(\text{reshape}(Z)), \quad \Psi_\phi: \mathbb{R}^{N \times d} \rightarrow \mathbb{R}^{H \times W}.
\label{eq:decoder_mapping}
\end{equation}

To supervise denoising, we minimize the $L1$ loss between the predicted and clean image:
\begin{equation}
\hat{I} = \arg\min_I \; \mathbb{E}_{I_{\text{clean}}} \left[ \left\| \Psi_\phi(\Phi_\theta(I_{\text{noisy}})) - I_{\text{clean}} \right\|_1 \right].
\label{eq:reconstruction_loss}
\end{equation}

\subsection{Prior-guided Hierarchical Instance–pixel
Contrastive Learning}
We propose a prior-guided hierarchical contrastive learning framework that jointly performs pixel- and instance-level contrastive learning to encourage noise-invariant and structure-aware representations.
Specifically, the framework consists of two complementary components: Statistics-Guided Pixel-Level Contrastive Learning (SPCL) and Memory-Augmented Instance-Level Contrastive Learning (MICL).


\subsubsection{Statistics-Guided Pixel-Level Contrastive Learning}
Statistics-Guided Pixel-Level Contrastive Learning (SPCL) is designed to enforce pixel-wise structural consistency by explicitly accounting for the statistical properties of speckle noise, thereby reducing noise-dominated interference while preserving fine-grained anatomical details. During the denoising process, the objective is to accurately distinguish noise from clean signal components and to restore the true pixel intensities from degraded input images. By leveraging pixel-level contrastive learning, the model can make more precise decisions regarding the status of each pixel.
In SPCL, positive samples are defined as the pixels at corresponding locations in the clean (ground truth) image, whereas negative samples are the pixel points generated in the noisy image under the guidance of prior knowledge. Through this method, the model gradually learns to distinguish clean pixels from noisy counterparts, thereby learning how to reduce or eliminate the impact of noise. 

Another significant advantage of SPCL lies in its enhanced ability to handle local noise and spatially varying patterns within an image. SPCL can capture local differences and variations, making the denoising process more refined and personalized. This capability is especially crucial when addressing non-uniform noise distributions or region-specific noise characteristics commonly observed in ultrasound imaging.

Our framework demonstrates the integration of the proposed SPCL module, which leverages feature information from the final decoder layer of both network branches. Let the output feature maps from the noisy and clean branches be denoted as $\mathbf{P}^N, \mathbf{P}^C \in \mathbb{R}^{N \times H \times W}$, respectively. We randomly select a spatial position $K = (i, j)$ in $\mathbf{P}^N$ as the \textit{anchor point} (denoted in red). Its corresponding position in $\mathbf{P}^C$ is defined as the \textit{positive sample} (denoted in green), i.e., $\mathbf{p}^+ = \mathbf{P}^C_{:, i, j}$. The anchor is thus $\mathbf{p}^a = \mathbf{P}^N_{:, i, j}$.

To capture local structural consistency, we randomly sample multiple such anchor points $\{K_k\}_{k=1}^M$ from $\mathbf{P}^N$, and for each $K_k$, define a square neighborhood $\Omega_k$ of size $s \times s$ (e.g., $s=3$ or $s=5$) centered at $K_k$. For each region $\Omega_k$, we compute the local mean and variance across all spatial positions and channels as:

\begin{equation}
\boldsymbol{\mu}_k = \frac{1}{|\Omega_k|} \sum_{(u,v) \in \Omega_k} \mathbf{P}^N_{:, u, v},
\end{equation}

\begin{equation}
\boldsymbol{\sigma}^2_k = \frac{1}{|\Omega_k|} \sum_{(u,v) \in \Omega_k} \left( \mathbf{P}^N_{:, u, v} - \boldsymbol{\mu}_k \right)^2
\end{equation}

These statistics act as prior-guided cues to enhance the network's sensitivity to local textures and structural variations, serving as auxiliary signals in the contrastive learning objective.

To identify severely corrupted regions dominated by speckle noise, we further compute the mean-to-standard-deviation ratio $\rho_K$ of each region:

\begin{equation}
\rho_K = \frac{\boldsymbol{\mu}_K}{\boldsymbol{\sigma}_K}
\end{equation}

Regions where $\rho_K < \tau_\text{noise}$ are considered to exhibit significant noise characteristics. Based on empirical observations across multiple ultrasound datasets, we set the threshold $\tau_\text{noise} = 1.92$. This threshold is supported by prior studies indicating that fully developed speckle noise in ultrasound images typically follows a Rayleigh distribution, where the signal-to-noise ratio (SNR) converges to approximately 1.91–1.92~\cite{destrempes2013segmentation, misra2013new}.

In our design, only regions satisfying this condition are selected as negative sample candidates for contrastive learning. These regions provide representative hard negatives that encourage the network to distinguish between noisy and clean spatial patterns in a more discriminative manner.

Accordingly, we define the pixel-level contrastive loss as:

\begin{equation}
\mathcal{L}_{\text{pix}} = - \sum_{K \in \mathcal{S}} \log \frac{\exp(\mathbf{p}^a_K \cdot \mathbf{p}^+_K / \tau)}{\exp(\mathbf{p}^a_K \cdot \mathbf{p}^+_K / \tau) + \sum_{j \in \mathcal{N}_K} \exp(\mathbf{p}^a_K \cdot \mathbf{p}^-_j / \tau)}
\end{equation}
where $\mathbf{p}^a_K$ and $\mathbf{p}^+_K$ denote the anchor and positive features at position $K$, and $\mathbf{p}^-_j$ represents negative features sampled from high-noise regions $\mathcal{N}_K$ as defined above. $\tau$ is a temperature scaling parameter.

\subsubsection{Memory-Augmented Instance-Level Contrastive Learning}
Memory-Augmented Instance-Level Contrastive Learning (MICL) is introduced to align global semantic representations between noisy and clean images in the global feature space. By contrasting image-level embeddings and leveraging a memory bank to provide a diverse set of negative samples, MICL mitigates speckle-induced semantic drift and promotes stable, noise-invariant instance representations.


In particular, Given a mini-batch of $B$ samples $\{(x_i^N, x_i^C)\}_{i=1}^{B}$, where $x_i^N$ is a noisy ultrasound image and $x_i^C$ is its corresponding clean counterpart, we extract their global feature embeddings using the shared-weight encoder $\mathcal{F}(\cdot)$:

\begin{equation}
\mathbf{z}_i^N = \mathcal{F}(x_i^N), \quad \mathbf{z}_i^C = \mathcal{F}(x_i^C)
\end{equation}
These embeddings $\mathbf{z}_i^N$ and $\mathbf{z}_i^C$ are treated as a positive pair, while embeddings from previous batches are used as negative samples. To ensure a diverse and memory-efficient negative sample pool, we adopt a first-in-first-out (FIFO) memory bank $\mathcal{M}$ with fixed capacity, which stores encoded feature vectors from prior mini-batches. Each feature $\mathbf{z}_i^N$ is encouraged to be close to its clean counterpart $\mathbf{z}_i^C$ (positive) while being dissimilar to the stored negative features $\mathbf{z}_j^{-} \in \mathcal{M}$. This contrastive objective is optimized using the InfoNCE loss:
\begin{equation}
\mathcal{L}_{\text{inst}} = - \sum_{i=1}^{B} \log \frac{\exp(\mathbf{z}_i^N \cdot \mathbf{z}_i^C / \tau)}{\exp(\mathbf{z}_i^N \cdot \mathbf{z}_i^C / \tau) + \sum_{\mathbf{z}_j^{-} \in \mathcal{M}} \exp(\mathbf{z}_i^N \cdot \mathbf{z}_j^{-} / \tau)}
\end{equation}
where $\tau$ denotes the temperature parameter controlling distribution sharpness.

After each training iteration, the current batch of clean features $\{\mathbf{z}_i^C\}_{i=1}^{B}$ is appended to $\mathcal{M}$, replacing the oldest entries when capacity is exceeded. 
This design enables the learning module to operate in the global feature space, explicitly aligning noisy and clean image representations and mitigating semantic drift caused by accumulated noise perturbations.

\subsection{The Hybrid Loss Function}
\textbf{Total loss:}
We employ a hybrid loss function composed of two components. The first component is the $L1$ loss, which quantifies the discrepancy between the predicted pixel values and the ground truth. The second component integrates contrastive learning loss, incorporating both prior knowledge-guided and instance-level loss functions. The formula is as follows:
\begin{equation} 
    \mathcal{L} = \mathcal{L}_{L1} + \mathcal{L}_{Con}
\end{equation}
where $\mathcal{L}_{L1}$ is $L1$ loss and $\mathcal{L}_{Con}$ is the hybrid contrastive loss.
\\
\indent \textit{\textbf{L1 loss:}} The $L1$ loss is used to calculate the difference between final output and the image without any noise added (ground truth). $L1$ loss between a clean image \( I_{clean} \) and a predicted image \( I_{pred} \) is given by:

\begin{equation}
    \mathcal{L}_{1} = \frac{1}{H \times W} \sum_{h=1}^{H} \sum_{w=1}^{W} \left| \mathbf{I}_{clean}(h,w) - \mathbf{I}_{pred}(h,w) \right|
\end{equation}
where \( H \) and \( W \) denote the height and width of the image, corresponding to the number of pixel rows and columns, respectively. \( \mathbf{I}_{clean}(h,w) \) and \( \mathbf{I}_{pred}(h,w) \) are the pixel value at position \( (h, w) \) in the clean image and predicted image respectively.
\\
\indent \textit{\textbf{Contrastive loss:}} 
The contrastive loss comprises two components: Statistics-Guided Pixel-Level Contrastive Learning and Memory-Augmented Instance-Level Contrastive Learning, which can be formulated as follows:
\begin{equation}
    \mathcal{L}_{Con} = \mathcal{L}_{Pixel} + \mathcal{L}_{Instance}
\end{equation}
where $\mathcal{L}_{\text{Pixel}}$ denotes the Statistics-Guided Pixel-Level Contrastive Learning loss, which enforces pixel-wise structural consistency by explicitly modeling the statistical characteristics of speckle noise. $\mathcal{L}_{Instance}$ denotes the Memory-Augmented Instance-Level Contrastive Learning Loss, aiming to align global semantic representations of noisy and clean images. Through image-level contrast and a memory bank that supplies diverse negatives, this module reduces speckle-induced semantic drift and stabilizes noise-invariant global representations.

Thus, the total loss function can be formulated as:
\begin{equation}
    \mathcal{L} = \mathcal{L}_{1} + \alpha \mathcal{L}_{Pixel} + \beta \mathcal{L}_{Instance}
\end{equation}
where $\alpha$ and $\beta$ are weighting parameters that balance the contributions of the pixel and instance level contrastive loss components to the total loss function.

\section{Experiment setting}
\subsection{Data Preprocessing}
We conduct experiments on two public ultrasound datasets, BUSI~\cite{al2020dataset} and CAMUS~\cite{leclerc2019deep}. As these datasets provide only clean images without corresponding noise labels, synthetic noise is introduced to generate noisy observations according to Eq. (1). Speckle noise is adopted as the noise model in this study, as it is a prevalent artifact in ultrasound imaging and has been extensively investigated. It is employed to reflect the characteristic noise behavior observed in practical ultrasound acquisitions. Speckle noise with varying intensities (0.25, 0.5, and 0.75) is applied to the ultrasound images to simulate noise conditions ranging from mild to severe. Through this procedure, a comprehensive dataset is constructed to facilitate the evaluation and comparison of different denoising techniques.

\subsection{Implementation Details}
All comparative and ablation experiments were implemented in PyTorch and executed on an NVIDIA GTX 4090 GPU. The resolution of all images was resized to $224 \times 224$ pixels. During the training phase, an initial learning rate of 0.001 was set, and a total of 200 training epochs were planned. To optimize the training efficiency, the AdamW optimizer was chosen, combined with the ReduceLROnPlateau strategy for dynamically adjusting the learning rate, with the aim of further accelerating training speed and improving the model's generalization capabilities. Furthermore, the training and testing sets were split with a ratio of 7:3 for both BUSI and CAMUS datasets. The specific quantities of the public datasets are shown in Table~\ref{dataset}. Based on the dataset split, the BUSI dataset comprises 546 images for training and 234 images for testing, whereas the CAMUS dataset consists of 1400 training images and 600 testing images.

\begin{table}[ht]
\centering
\caption{Division of Dataset. Summary of the dataset partitioning for training, testing and total samples in the BUSI and CAMUS datasets.}
\begin{tabular}{cccc}
\toprule
\textbf{Dataset} & \textbf{Train} & \textbf{Val} & \textbf{Total $(\#Num)$} \\ 
\midrule
BUSI                & 546            & 234            & 780          \\ \midrule
CAMUS               & 1400           & 600            & 2000         \\ 
\bottomrule
\end{tabular}
\label{dataset}
\end{table}

\subsection{Evaluation Metrics}
PSNR (Peak Signal-to-Noise Ratio) and SSIM (Structural Similarity Index)~\cite{hore2010image} are commonly used to evaluate image quality in denoising task. In addition, RMSE (Root Mean Square Error) \cite{hodson2022root} is employed as an evaluation metric to further assess denoising performance.
\subsubsection{SSIM} It measures the structural similarity between the original and denoised images and is more consistent with human visual perception.
\begin{equation}
SSIM(x, y) = \frac{(2\mu_x \mu_y + C_1)(2\sigma_{xy} + C_2)}{(\mu_x^2 + \mu_y^2 + C_1)(\sigma_x^2 + \sigma_y^2 + C_2)}
\end{equation}

\subsubsection{PSNR} It measures the ratio between the maximum possible signal power and the power of the corrupting noise, providing an objective measure of reconstruction fidelity.
\begin{equation}
PSNR = 10 \log_{10} \left( \frac{MAX^2}{MSE} \right)
\end{equation}
\subsubsection{RMSE}
RMSE quantifies the pixel-wise difference between the predicted and original images, providing a straightforward measure of restoration error.

\begin{equation}
RMSE = \sqrt{\frac{1}{w \times h} \sum_{i=1}^{w} \sum_{j=1}^{h} \left[ I_{ori}(i, j) - I_{pred}(i, j) \right]^2}
\end{equation}

\section{Results and discussion}

\begin{table*}[t]

\caption{
We present a comparative analysis of denoising results on both BUSI and CAMUS datasets. Experimental evaluations were carried out at noise levels of 0.25, 0.5 and 0.75 respectively. The best results across these evaluations are highlighted in bold for clarity and emphasis.
}
\label{table:bss_sdr}\vspace{-0.2cm}
\centering
\resizebox{0.9\textwidth}{!}{
\begin{tabular}{ c c c c c c c c c c c } 
\toprule
\multirow{2}{*}{Datasets} & \multirow{2}{*}{Methods} & \multicolumn{3}{c}{$\sigma$ = 0.25} & \multicolumn{3}{c}{$\sigma$ = 0.5} & \multicolumn{3}{c}{$\sigma$ = 0.75} \\

\cmidrule(lr){3-5} \cmidrule(lr){6-8} \cmidrule(lr){9-11} 
 &  & PSNR & SSIM & RMSE & PSNR & SSIM & RMSE & PSNR & SSIM & RMSE \\

\midrule

& BM3D\cite{dabov2007image}        & 27.9196 & 0.8264 & 7.6514 & 28.6314 & 0.8139 & 7.6498 & 27.6144 & 0.8065 & 7.9549 \\ 
& UNet\cite{ronneberger2015u}      & 38.1121 & 0.9816 & 3.0182 & 34.4947 & 0.9577 & 4.2414 & 33.4131 & 0.9405 & 4.4913 \\ 
& DnCNN\cite{zhang2017beyond}      & 36.7261 & 0.9773 & 3.2945 & 34.6723 & 0.9591 & 4.0224 & 33.1491 & 0.9360 & 4.5809 \\
CAMUS & RED-CNN\cite{chen2017low}  & 29.8887 & 0.8778 & 5.4783 & 29.2920 & 0.9029 & 5.3683 & 28.1246 & 0.8724 & 5.8272 \\ 
& SwinIR\cite{liang2021swinir}     & 34.5439 & 0.9689 & 4.4822 & 34.1665 & 0.9556 & 4.3323 & 33.7896 & 0.9429 & 4.4731 \\ 
& Uformer\cite{wang2022uformer}    & 38.5574 & 0.9815 & 2.9227 & 34.5160 & 0.9591 & 4.1921 & 33.5542 & 0.9389 & 4.4211 \\ 
& Restormer \cite{zamir2022restormer}& 38.4462 & 0.9805 & 3.1848 & 34.5170 & 0.9584 & 4.1875 & 33.5354 & 0.9385 & 4.4654\\
\rowcolor{gray!15}
& \textcolor{red}{\textbf{Proposed}} & \textbf{38.8399} & \textbf{0.9842} & \textbf{2.8928} & \textbf{34.7329} & \textbf{0.9612} & \textbf{4.0132} & \textbf{33.8249} & \textbf{0.9435} & \textbf{4.3495} \\

\midrule
 & BM3D\cite{dabov2007image}      & 22.4846 & 0.8165 & 9.6218 & 21.5155 & 0.8004 & 9.6314 & 22.3558 & 0.8057 & 9.1061 \\ 
 & UNet\cite{ronneberger2015u}    & 22.0179 & 0.8345 & 9.4409 & 22.0002 & 0.8048 & 9.4249 & 24.0726 & 0.8080 & 8.8002 \\
& DnCNN\cite{zhang2017beyond}     & 25.9552 & 0.8342 & 7.8082 & 24.1854 & 0.7887 & 8.1447 & 24.3415 & 0.7655 & 8.1674 \\ 
BUSI & RED-CNN\cite{chen2017low}  & 32.4131 & 0.9273 & 4.9464 & 26.4941 & 0.7901 & 8.0179 & 25.3338 & 0.7619 & 7.7326 \\
& SwinIR\cite{liang2021swinir}    & 35.7962 & 0.9457 & 3.8637 & 32.5459 & 0.9428 & 4.5876 & 31.8615 & 0.9141 & 5.2568 \\ 
& Uformer\cite{wang2022uformer}   & 36.4134 & 0.9689 & 3.4251 & 32.4754 & 0.9431 & 5.1203 & 31.2750 & 0.9153 & 5.5109 \\
& Restormer\cite{zamir2022restormer} & 36.4244 & 0.9687 & 3.6884 & 32.4852 & 0.9428 & 5.1782 & 31.2595 & 0.9122 & 5.6498 \\
\rowcolor{gray!15}
& \textcolor{red}{\textbf{Proposed}}  & \textbf{36.5392} & \textbf{0.9703} & \textbf{3.3214} & \textbf{32.6842} & \textbf{0.9472} & \textbf{4.3251} & \textbf{32.1351} & \textbf{0.9173} & \textbf{4.8491} \\ 

\bottomrule
\end{tabular}
}
\label{table2}

\end{table*}

\begin{figure*}[!t]
\centerline{\includegraphics[width=0.9\textwidth]{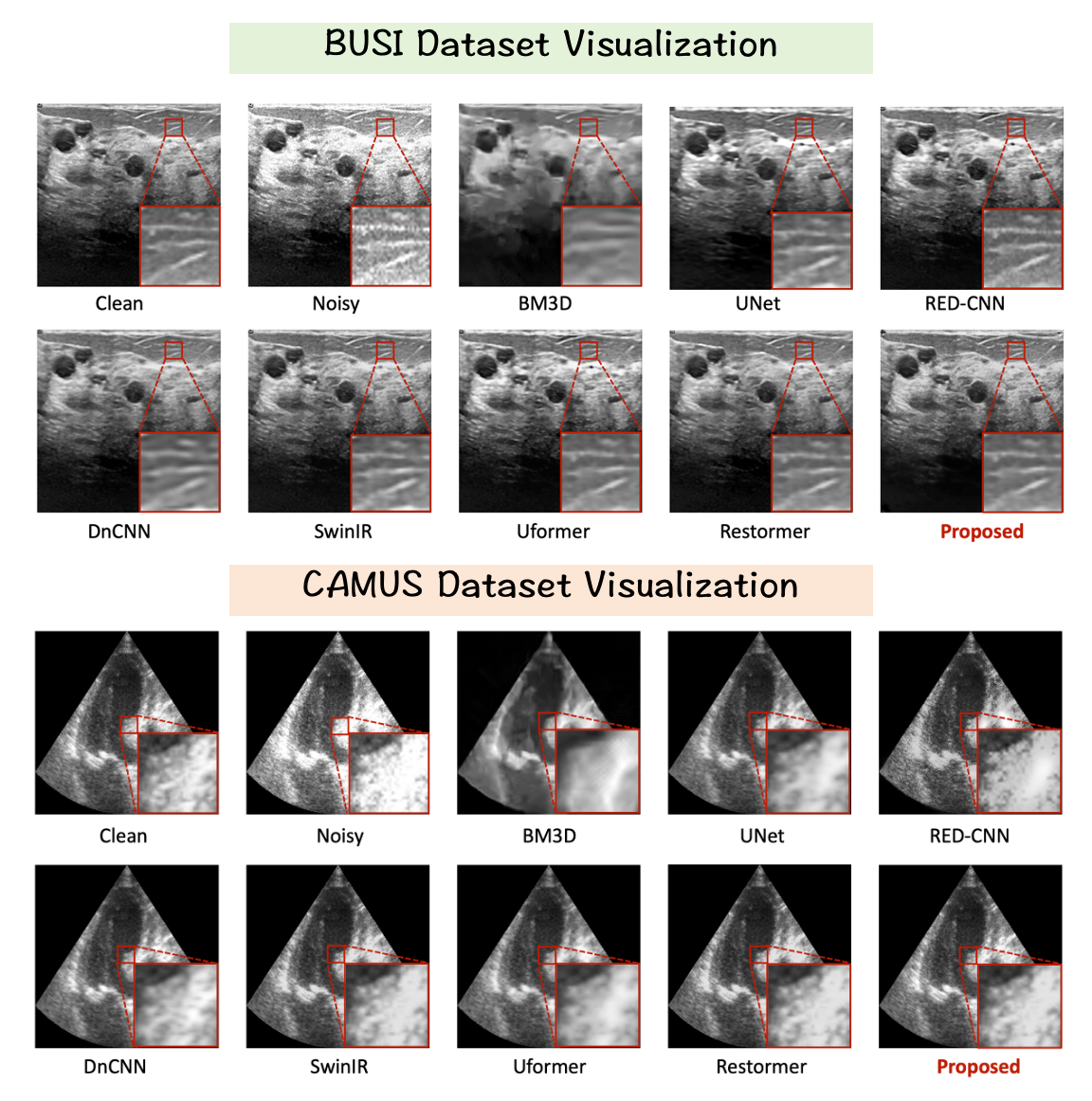}}
\caption{Qualitative comparison of denoising results on representative ultrasound images (CAMUS and BUSI). The top row in each subplot displays the clean reference, noisy input, and outputs from classical and deep learning-based baselines (BM3D, UNet and RED-CNN). The bottom row shows results obtained under DnCNN, SwinIR, Uformer, Restormer, and our proposed method. Red boxes indicate regions of interest for detailed visual inspection.}
\label{fig1}
\end{figure*}


\subsection{Comparison experiments}
We conduct the comparative experiments, aiming to benchmark the proposed method against the performance of state-of-the-art image restoration or denoising models. The comparison methods included three categories: classical denoising techniques, such as BM3D, deep learning methods based purely on convolutional neural networks (CNNs), which have gained widespread used in recent years for their excellent feature extraction capabilities and methods based on Transformers, which leverage self-attention mechanisms to capture long-range dependencies in images. We show the comparative experimental results on two public datasets, BUSI and CAMUS datasets respectively. Figure~\ref{fig1} shows the visualization results based on two representative images.

From Table~\ref{table2}, we report the performance of various denoising methods on the BUSI and CAMUS datasets. $\sigma$ represents different levels of noise standard deviation, specifically 0.25, 0.5 and 0.75 in our experiment setting. Larger values correspond to higher noise levels, resulting in more severe noise corruption across the images. PSNR, SSIM, and RMSE are employed to evaluate denoising performance, where higher PSNR and SSIM values and lower RMSE values indicate better image quality and more effective noise suppression.
For the CAMUS dataset, BM3D achieves modest denoising performance, while certain CNN-based methods, including RED-CNN, demonstrate limited effectiveness. At \( \sigma = 0.5 \), BM3D and RED-CNN obtain SSIM values of 0.8139 and 0.9029, respectively, which are comparatively lower than those achieved by other methods.
Among this table, CNN-based models show moderate performance. UNet and DnCNN achieve performance comparable to that of transformer-based models. Among CNN-based methods, DnCNN attains the highest SSIM of 0.9591, while Uformer achieves the highest SSIM among transformer-based methods with the same value.
However, Uformer attains a slightly lower PSNR than DnCNN, with values of 34.5610 and 34.6723, respectively.
By leveraging the proposed prior-guided hierarchical contrastive learning framework, our model achieves superior performance, reaching peak values of 34.7329 in PSNR and 0.9612 in SSIM. At a noise level of \( \sigma = 0.75 \), severe noise is applied to the ultrasound images. Under this setting, BM3D and RED-CNN exhibit comparatively lower PSNR and SSIM values, whereas the proposed method achieves the best overall performance.
\par
Experimental results on the BUSI dataset indicate that the transformer-based models generally outperform the CNN-based methods. One possible explanation for this performance difference is the variation in image resolution, as BUSI images generally have higher spatial resolution than those in the CAMUS dataset. This performance difference may be attributed to the ability of transformer-based models to capture long-range spatial dependencies, which become increasingly important in high-resolution images. In contrast, CNN-based methods primarily rely on local receptive fields, which may be insufficient for modeling large-scale structural consistency in high-resolution ultrasound images. \par
However, the integration of both remains complementary. For low-resolution images, CNN-based models may be less effective in exploiting their local receptive field advantages, as the available contextual information is limited and local features alone may be insufficient for comprehensive denoising. In contrast, Transformer-based models are designed to capture global contextual information, allowing them to handle images effectively and deliver superior denoising results even for smaller images. Under high-noise conditions, our proposed method consistently improves image quality, demonstrating strong robustness and adaptability across different datasets. At a noise level of $\sigma = 0.75$, all CNN-based methods including BM3D, yield PSNR values below 30. In contrast, Transformer-based models such as SwinIR, Uformer and Restormer surpass a PSNR of 31, highlighting the strength of global feature modeling. On this dataset, the proposed method achieves the highest performance, benefiting from the prior-guided hierarchical instance–pixel contrastive learning strategy and the hybrid network.

Overall, the proposed method effectively learns noise-invariant and structure-aware representations through the prior-guided hierarchical instance–pixel contrastive learning framework along with the Transformer-CNN architecture, leading to superior denoising performance across two public ultrasound datasets.

\subsection{Ablation study}
In this section, we conduct an ablation study to evaluate the effectiveness of the proposed components. Here we implement our ablation study under BUSI dataset with $\sigma=0.5$. Table \ref{results_encoder_decoder_order} presents the results of the ablation experiments for the three parts (Transformer-CNN Model, Statistics-Guided Pixel-Level Contrastive Learning and Memory-Augmented Instance-Level Contrastive Learning. This table compares the performance of ultrasound image denoising models under four different configurations. Each column's \Checkmark indicates that the corresponding part is enabled in the model, while \XSolidBrush indicates that the part is not included. Here, PSNR, SSIM and RMSE are selected as metrics to evaluate the denoising performance of the models.

From the perspective of PSNR, the model with all part enabled achieved a PSNR value of 32.6842, indicating that the model can restore images with high accuracy. Compared to other configurations, the complete model demonstrates a significant improvement in retaining image details and structure. Regarding SSIM, which reflects the amount of structural information retained after image denoising, the complete model also exhibits the best performance, reaching 0.9472. This result indicates that the model not only enhances the visual quality of images but also better preserves their structural details under our specific design. For RMSE, which measures the pixel-level error between the original image and the denoised image, the complete model achieves the lowest RMSE value of 4.3251, signifying the highest prediction accuracy at the pixel level.

When either Statistics-Guided Pixel-Level Contrastive Learning and Memory-Augmented Instance-Level Contrastive Learning are removed, declines in PSNR and SSIM are observed and RMSE increases. This demonstrates that two types of contrastive learning contribute to the overall performance improvement of the model. The most critical observation is that when the hybrid architecture (i.e., Hybrid, Pixel Level, and Instance Level) is entirely omitted, all performance metrics deteriorate significantly: PSNR drops to 22.0002, SSIM decreases to 0.8048 and RMSE rises to 9.4249.

These results indicate that both the prior-guided hierarchical contrastive learning strategy and the hybrid architecture contribute to improved denoising performance.


\begin{table}[ht]
\centering
\caption{Ablation experiments evaluating the impact of hybrid, pixel-level, and instance-level modules on PSNR, SSIM, and RMSE. Results are based on the BUSI dataset with a noise level of 0.5.}
\label{tab:ablationModule}
\setlength{\tabcolsep}{4.8pt} 
\renewcommand{\arraystretch}{1.2} 

\resizebox{0.9\linewidth}{!}{ 
\begin{tabular}{ccc|ccc}
\toprule
\multicolumn{3}{c|}{\textbf{Module}} & \multicolumn{3}{c}{\textbf{Metrics}} \\
\hline
\textbf{Hybrid} & \textbf{Pixel} & \textbf{Instance} & \textbf{PSNR↑} & \textbf{SSIM↑} & \textbf{RMSE↓} \\
\midrule
\XSolidBrush & \XSolidBrush & \XSolidBrush & 22.0002 & 0.8048 & 9.4249 \\
\Checkmark   & \XSolidBrush & \XSolidBrush & 32.1449 & 0.9349 & 5.0192 \\
\Checkmark   & \Checkmark   & \XSolidBrush & 32.4251 & 0.9423 & 4.5154 \\ 
\rowcolor{gray!15}
\Checkmark   & \Checkmark   & \Checkmark   & \textcolor{red}{32.6842}& \textcolor{red}{0.9472} & \textcolor{red}{4.3251} \\
\bottomrule
\end{tabular}
}
\vspace{-2mm}
\end{table}

\begin{table}[ht]
\centering
\caption{Ablation experiments to evaluate the impact of encoder-decoder orders on PSNR, SSIM, and RMSE metrics.}
\label{results_encoder_decoder_order}
\setlength{\tabcolsep}{4.8pt} 
\renewcommand{\arraystretch}{1.2} 

\resizebox{0.9\linewidth}{!}{
\begin{tabular}{ccccc}
\toprule

\multicolumn{2}{c}{\textbf{Order}} & \multicolumn{3}{c}{\textbf{Metrics}} \\
\hline
\textbf{Encoder} & \textbf{Decoder} & \textbf{PSNR↑} & \textbf{SSIM↑} & \textbf{RMSE↓} \\ 
\midrule
Transformer & Transformer & 32.1421 & 0.9325 & 4.9828 \\
CNN         & Transformer & 32.5221 & 0.9312 & 5.0252 \\
CNN         & CNN         & 22.0002 & 0.8048 & 9.4249 \\
\rowcolor{gray!15}
Transformer & CNN         & \textcolor{red}{32.6842} & \textcolor{red}{0.9472} & \textcolor{red}{4.3251} \\ \bottomrule
\end{tabular}
}
\vspace{-2mm}
\end{table}

\subsection{Hyper-parameter experiments}
We further investigate the impact of key hyper-parameters on model performance, including the loss weight coefficients and the depth of the transformer blocks in the encoder.

\subsubsection{Weight of loss function}



\begin{table}[ht]
    \centering
    \caption{Hyperparameter analysis of the pixel level loss weight and its impact on model performance.}

    \setlength{\tabcolsep}{4.8pt} 
    \renewcommand{\arraystretch}{1.2} 
    
    \resizebox{0.7\linewidth}{!}{ 
    \begin{tabular}{cccc}
        \toprule
        $\alpha_1$ & \textbf{PSNR}~$\uparrow$ & \textbf{SSIM}~$\uparrow$ & \textbf{RMSE}~$\downarrow$ \\
        \midrule
        0.1 & 34.5481 & 0.9578 & 4.3264 \\
        0.3 & 34.4264 & 0.9584 & 4.2185 \\
        \rowcolor{gray!15}
        0.5 & \textcolor{red}{34.7194} & \textcolor{red}{0.9603} & \textcolor{red}{4.1154} \\
        0.7 & 34.6062 & 0.9589 & 4.2558 \\ 
        \bottomrule
    \end{tabular}
    }
    \label{hyperparam_alpha_one}
\end{table}

\begin{table}[ht]
    \centering
    \caption{Hyperparameter analysis of the instance level loss weight and its impact on model performance.}
    \setlength{\tabcolsep}{4.8pt} 
    \renewcommand{\arraystretch}{1.2} 
    
    \resizebox{0.8\linewidth}{!}{
    \begin{tabular}{ccccc}
        \toprule
        $\alpha_1$ & $\alpha_2$ & \textbf{PSNR}~$\uparrow$ & \textbf{SSIM}~$\uparrow$ & \textbf{RMSE}~$\downarrow$ \\
        \midrule
        0.5 & 0.1 & 34.6621 & 0.9591 & 4.2511 \\
        0.5 & 0.3 & 34.7051 & 0.9597 & 4.0823 \\
        0.5 & 0.5 & 34.6721 & 0.9587 & 4.2648 \\
        \rowcolor{gray!15}
        0.5 & 0.7 & \textcolor{red}{34.7329} & \textcolor{red}{0.9612} & \textcolor{red}{4.0132} \\
        \bottomrule
    \end{tabular}
    }
    \label{hyperparam_alpha_two}
\end{table}

We further analyze the effect of different loss weight coefficients.
Table ~\ref{hyperparam_alpha_one} explores the impact of different weights for Statistics-Guided Pixel-Level Contrastive Learning loss function on the experimental results, with weights of \( \alpha_1 = 0.1, 0.3, 0.5, 0.7 \) being tested. From the final results, the best performance is achieved when \( \alpha_1 = 0.5 \).

Given \( \alpha_1 = 0.5 \), Table~\ref{hyperparam_alpha_two} investigates the effect of the weight for Memory-Augmented Instance-Level Contrastive Learning loss function based on high-dimensional features. Under the condition \( \alpha_1 = 0.5 \), experiments were conducted with \( \alpha_2 = 0.1, 0.3, 0.5, 0.7 \). As shown in Table~\ref{hyperparam_alpha_two}, the optimal performance is obtained when \( \alpha_1 = 0.5 \) and \( \alpha_2 = 0.7 \).

\subsubsection{Depth of transformer blocks}

\begin{table}[ht]
    \centering
    \caption{Performance metrics under different transformer block depth configurations.}
    \setlength{\tabcolsep}{4.8pt} 
    \renewcommand{\arraystretch}{1.2} 
    
    \resizebox{0.8\linewidth}{!}{ 
    \begin{tabular}{lccc}
        \toprule
        \textbf{Depth} & \textbf{PSNR}~$\uparrow$ & \textbf{SSIM}~$\uparrow$ & \textbf{RMSE}~$\downarrow$ \\
        \midrule
        2, 2, 32, 2 & 30.6842 & 0.9158 & 5.0515 \\
        \rowcolor{gray!15}
        2, 4, 32, 2 & \textcolor{red}{32.6842} & \textcolor{red}{0.9472} & \textcolor{red}{4.3251} \\
        2, 8, 32, 2 & 31.5149 & 0.9175 & 5.0195 \\
        2, 16, 32, 2 & 30.8489 & 0.9251 & 4.8942 \\
        \bottomrule
    \end{tabular}
    }
    
    \label{depth_comparison}
\end{table}
Table~\ref{depth_comparison} illustrates the impact of different transformer block depths on image denoising performance. Here, depth refers to the number of stacked layers within the Transformer network. Several transformer depth configurations are experimentally set and evaluated, including (2, 2, 32, 2), (2, 4, 32, 2), (2, 8, 32, 2), and (2, 16, 32, 2). We observe that the configuration with a depth of 2, 4, 32, 2 achieved optimal results in terms of both PSNR and SSIM, indicating that this depth strikes a balance between performance and computational complexity. Additionally, the RMSE metric reached its lowest value under this configuration, reflecting the smallest error between the predicted and ground truth images. 

As the depth increases, slight improvements in PSNR and SSIM are observed for some configurations; however, the gains are marginal and the reduction in RMSE remains limited. This suggests that increasing the depth beyond a certain point does not yield substantial performance gains and may increase the risk of overfitting.

\section{Conclusion}
In this paper, we introduce a prior-guided hierarchical instance–pixel contrastive learning framework that promotes noise-robust and structure-aware representations by increasing the distinction between noisy and clean samples at both pixel and instance levels. Statistics-Guided Pixel-Level Contrastive Learning module leverages prior statistical cues to enhance local structural consistency, while Memory-Augmented Instance-Level Contrastive Learning module aligns global representations between noisy and clean images, mitigating semantic drift induced by noise perturbations. Coupled with a Transformer–CNN backbone that integrates global context modeling and local detail preservation, our proposed method demonstrates robust denoising performance across different noise levels on two public ultrasound datasets.



\section*{References}
\bibliographystyle{ieeetr} 
\bibliography{bibliography}

\end{document}